# One Interface, Many Robots: Unified Real-Time Low-Level Motion Planning for Collaborative Arms

Yue Feng, Weicheng Huang, and I-Ming Chen, *Fellow, IEEE*

*Abstract* — This paper proposes a common interface for real-time low-level motion planning of collaborative robotic arms, aimed at enabling broader applicability and improved portability across heterogeneous hardware platforms. In previous work, we introduced WinGs Operating Studio (WOS), a middleware solution that abstracts diverse robotic components into uniform software resources and provides a broad suite of language-agnostic APIs. This paper specifically focuses on its minimal yet flexible interface for real-time end-effector trajectory control. By employing an n-degree polynomial interpolator in conjunction with a quadratic programming solver, the proposed method generates smooth, continuously differentiable trajectories with precise position, velocity, and acceleration profiles. We validate our approach in three distinct scenarios. First, in an offline demonstration, a collaborative arm accurately draws various geometric shapes on paper. Second, in an interruptible, low-frequency re-planning setting, a robotic manipulator grasps a dynamic object placed on a moving mobile robot. Finally, we conducted a teleoperation experiment in which one robotic arm controlled another to perform a series of dexterous manipulations, confirming the proposed method's reliability, versatility, and ease of use.

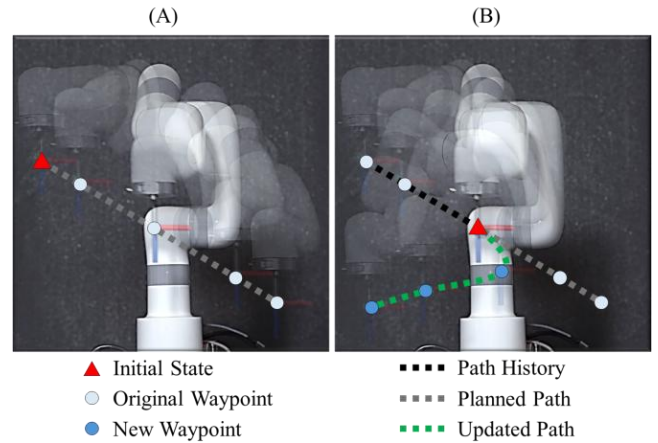

Fig. 1.  Motion Planning Visualization. A) Planned Cartesian motion of the robotic arm end-effector on the x–z plane through four waypoints, with uniform 2-second durations. Motion trails illustrate key snapshots. B) Online replanning triggered at the red triangle introduces three new waypoints. The arm smoothly transitions to the updated path and lands on the lower-left corner on time.

## I. INTRODUCTION

COLLABORATIVE robotic arms have become increasingly prevalent in a wide range of applications, spanning industrial automation, commercial services, and domain-specific tasks. In industrial environments, tasks such as welding [1], painting [2], and palletizing [3] have been automated through collaborative robot systems. Similarly, commercial applications, including robotic baristas [4], massage robots [5], and robotic waiters [6], have begun to integrate into our everyday lives. Moreover, teleoperation technologies have emerged as a key solution for addressing challenges in hazardous or extreme environments, such as chemical processing [7], nuclear waste handling [8], and underwater manipulation [9].

Despite recent advances, significant barriers persist in the development and deployment of collaborative robot applications, due to the complexity of multidisciplinary system design, hardware heterogeneity, and the absence of standardized software interfaces. These issues result in steep learning curves, integration difficulties, and limited software portability across platforms. To address these persistent challenges, we previously introduced WinGs Operating Studio (WOS) [10], a robotic middleware that abstracts diverse robotic hardware into uniform software resources and provides intuitive, language-agnostic APIs. While such middleware [10–12] simplifies system integration, mechanism-independent low-level motion planning remains a significant challenge, particularly in some of the abovementioned applications where online trajectory replanning or real-time robust Cartesian-space teleoperation is required during execution.

Existing solutions are often tightly coupled with specific hardware, rely on proprietary interfaces [13-14]. As a result, high-level motion planning made for specific applications are difficult to reuse across platforms, and modifying trajectory execution logic often demands extensive, hardware-specific adjustments.

Widely adopted frameworks such as ROS MoveIt [15] have been instrumental in simplifying inverse kinematics (IK) computation and path planning across various robotic platforms [16], [17]. Nevertheless, these mainstream tools have limitations in real-time responsiveness and interruptibility but focus on optimizing offline motion plan. MoveIt typically relies on sampling-based planners, such as OMPL [18], which, although general-purpose and versatile, are inherently less predictable in highly dynamic environments or when strict real-time constraints are required.

Therefore, in this paper, we propose and implement a Unified

Yue Feng and I-Ming Chen are with Robotics Research Center (RRC), Nanyang Technological University, Singapore, 639798. (e-mail: yue011@e.ntu.edu.sg; michen@ieee.org)

Weicheng Huang is with WinGs Robotics LLC, NY, 10314 USA. (e-mail: info@wingsrobotics.com)

Real-Time Low-Level Motion Planning API integrated into the WOS middleware. Our proposed interface provides a straightforward yet powerful method for generating smooth, continuously differentiable trajectories in joint space, specified through Cartesian waypoints, capable of controlling diverse robotic arm hardware that supported by WOS.

Our approach draws inspiration from the literature on optimal motion planning [19–23], which has explored a variety of methods balancing trajectory smoothness and dynamic feasibility. Building on these ideas, we designed a solution that first transforms Cartesian waypoints into joint space using a generic inverse kinematics (IK) solver [24] and then applies n-degree polynomial interpolation combined with a quadratic programming (QP) solver to generate joint trajectories with precise position, velocity, and acceleration profiles. This combination ensures both real-time performance and compliance with the physical constraints of diverse robotic platforms.

The developed function supports a variety of operational scenarios, including offline trajectory execution, interruptible online replanning (as illustrated in Figure 1), and high-frequency real-time control. These capabilities address the needs of structured environments tasks, semi-structured collaborative settings, and robust teleoperation for dexterous manipulation.

The key contributions of this paper are as follows:
- *Motion Planning Interface:* A lightweight and versatile motion planning interface based on time-parameterized Cartesian waypoints is developed to enable consistent motion control across multiple collaborative robotic arms supported by WOS.
- *Trajectory Interpolation Method:* An interpolation method is formulated using n-degree polynomial curves in combination with quadratic programming optimization, which ensures trajectory smoothness while satisfying strict per-cycle computational constraints.
- *Evaluation Across Representative Tasks:* The effectiveness of the proposed interface is demonstrated through three distinct use cases: offline trajectory execution, interruptible dynamic grasping, and real-time Cartesian-space teleoperation between robotic arms.

The remainder of the paper is organized as follows: Section II introduces the motion planning interface and reviews relevant components of the WOS framework. Section III presents the algorithmic foundations. Section IV describes experimental validation and results across the three use cases. Section V concludes with a summary and directions for future work.

## II. INTERFACE DESIGN

This section presents the design of a unified interface that abstracts the complexities of robotic control into modular, reusable components. By leveraging WOS's node-based architecture, it enables consistent end-effector manipulation and motion planning across heterogeneous robotic platforms.

### A. Overview of the WOS Framework

WOS [10] is a lightweight and extensible robot operating platform conceptualized as a Robot as a Service (RaaS) system [25]. Different from well-known frameworks such as ROS [11], WOS is designed for rapid prototyping and deployment of robotics applications across diverse hardware platforms, including PCs, embedded systems, edge devices, and Android mobile devices. By distributing WOS as an executable binary, it eliminates the need for complex dependency management and enhances runtime security. Additionally, a web-based graphical programming interface is provided for users without coding experience, thereby lowering the entry barrier for robot application development [26-27].

In WOS, robotic hardware is abstracted into virtual Components, each consisting of a Driver and a Handler. The Driver handles low-level hardware communication, while the Handler manages logic execution. These Components are encapsulated as Nodes and are accessible through standardized Services. In addition to hardware-related components, Nodes can also encapsulate standalone algorithmic or application-level functionalities such as kinematics solvers, motion planning module, and vision-based object recognition module. These functional nodes integrate into the WOS ecosystem and collaborate with other nodes to form complete robotic applications.

WOS defines three core service interaction models—Topics, Requests, and Actions—to facilitate modular and scalable communication. These services enable event-driven messaging, synchronous operations, and long-running task execution with feedback and cancellation mechanisms. Both components and functional nodes can be organized hierarchically, enabling complex robotic structures such as manipulators and end-effectors to be integrated and managed alongside flexible application-level logic.

To ensure maximum flexibility and interoperability, WOS decouples the transport layer from its application logic and exposes all APIs through widely supported network protocols, including HTTP, WebSocket, WebRTC, MQTT, and TCP. This architectural choice allows WOS services to be accessed from any programming language capable of network communication, such as Python, JavaScript, or C++. As a result, developers can integrate and control WOS-based robotic systems using their preferred development environments without the need for WOS-specific dependencies. This design enhances the accessibility and portability of robotic applications across diverse development ecosystems [28-31].

### B. Motion Planning API

In many real-world robotic applications, Cartesian-space control of the end-effector is often more critical than direct joint-space manipulation. Numerous studies focus exclusively on the spatial position of the gripper, such as in grasp affordance learning from visual cues or in the development of sim-to-real reinforcement learning environments [32–33]. From a software architecture perspective, this abstraction implies that when only the end-effector's Cartesian position is of concern, the underlying robot model can be decoupled from the application layer. As long as the hardware satisfies essential physical constraints (e.g., workspace, payload), the same high-level control logic can be reused across different robot platforms. This motivates the development of a unified motion control framework that enables deployment once the robot configuration is properly defined.

To support this abstraction, WOS encapsulates key low-level functionalities, such as IK solvers, motion planning modules, PID controllers, and hardware drivers, into internal system Nodes. These Nodes are registered within WOS and designed for high-frequency, high-bandwidth communication to ensure soft real-time execution. Together, they form a modular, robot-agnostic control pipeline capable of driving a wide range of collaborative robots in Cartesian space.

As illustrated in Figure 2, WOS enables a unified real-time end-effector control pipeline across heterogeneous collaborative arms in Cartesian space. Each robot (e.g., Robot A, B, C) is preconfigured in the system with a set of parameters such as joint limits, link lengths, control frequency, and the transformation between the end-effector and the last joint.

Communication with robot hardware is handled by the WOS hardware driver, which supports a variety of low-level communication protocols. For robots that provide only torque-level interfaces (e.g., Robot C), WOS integrates a PID controller to enable accurate joint-position tracking, thereby standardizing the interface to a joint-position-controllable model. If velocity or acceleration interfaces are natively supported by hardware, the control performance can be further enhanced.

Upon receiving an external request, such as the *rt-move-cartesian* command introduced in this paper, the low-level motion planner delegates inverse kinematics computation to the IK solver Node and subsequently performs trajectory interpolation based on the returned joint-space targets. Each request contains a robot ID, a request type, and a structured payload. The planner converts this information into a stream of joint-space reference states (including position, velocity, and acceleration), which are published to the hardware driver at the configured control frequency.

The execution pipeline consists of two main threads. One thread continuously listens for incoming motion commands, performing real-time updates to the motion plan by invoking the IK solver and trajectory interpolation modules. The other thread operates at a fixed control frequency, dispatching the most recent reference points to the robot hardware. The *rt-move-cartesian* interface supports asynchronous and interruptible requests and is robust to non-uniform transmission rates. The interpolation process ensures that the resulting trajectory is continuously differentiable in position, velocity, and acceleration, while minimizing jerk to guarantee smooth motion and dynamic feasibility.

Users only need to provide a list of waypoints, where each waypoint specifies a target end-effector pose and a desired arrival time. Velocity and acceleration profiles are automatically generated by the motion planner. The initial pose is inferred from the robot's current state, and the final waypoint is constrained to zero velocity and acceleration to ensure a smooth stop at the end of the motion.

This API can be invoked from any programming language over standard network protocols supported by WOS, allowing flexible integration with external systems. Regardless of the underlying robotic hardware, WOS ensures transparent execution of Cartesian trajectories.

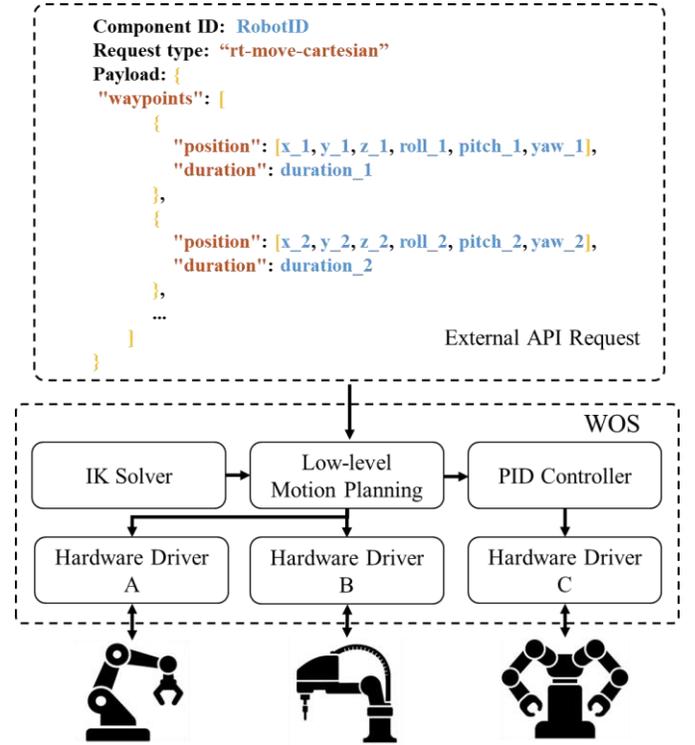

**Fig. 2.** Real-time end-effector control pipeline in WOS for heterogeneous collaborative robots.

## III. MOTION PLANNING ALGORITHM

The mathematical formulation of the trajectory interpolation and execution scheduling is detailed in this section, providing the algorithmic foundation for the unified Cartesian end-effector control framework introduced in Section II.

### A. Problem Formulation

We define the input Cartesian waypoint list as
$$\mathcal{C} = \{\xi_1, \xi_2, \dots, \xi_N\},$$
where each waypoint $\xi_i$ is defined as
$$\xi_i = (\mathbf{x}_i, D_i), \ \mathbf{x}_i \in \mathbb{R}^6, \ i = 1, \dots, N.$$
For a 6-DOF arm, the Cartesian pose $\mathbf{x}_i$ can be written as
$$\mathbf{x}_i = [x_i, y_i, z_i, \phi_i, \theta_i, \psi_i]^T,$$
where $x_i, y_i, z_i$ denote the translation of the end-effector with respect to the base frame, and $\phi_i, \theta_i, \psi_i$ are Euler angles representing orientation. The rotation convention of the Euler angles follows the specification required by the IK solver implementation.

Each $D_i$ denotes the time duration to reach $\mathbf{x}_i$ from $\mathbf{x}_{i-1}$, and the cumulative time stamp is denoted as
$$T_i = \sum_{i=1}^{N} D_i$$

Similarly, we define the joint-space waypoint list as
$$\mathcal{J} = \{\mathbf{j}_1, \mathbf{j}_2, \dots, \mathbf{j}_N\},$$
where each waypoint $\mathbf{j}_i$ is defined as
$$\mathbf{j}_i = (\mathbf{d}_i, D_i), \ \mathbf{d}_i \in \mathbb{R}^{dof}, \ i = 1, \dots, N.$$
Here, $\mathbf{d}_i$ is a joint position vector with a length equal to the number of joints in the manipulator. It represents the joint configuration required to realize the Cartesian pose. The joint-space waypoints are generated from the Cartesian sequence via inverse kinematics

as follows:
$$\mathbf{d}_i = \mathcal{K}^{-1}(\mathbf{x}_i; \mathbf{d}_{i-1}) \quad (1).$$
where $\mathcal{K}^{-1}(\cdot)$ denotes the IK solver and $\mathbf{d}_{i-1}$ is used as the reference or initial guess to ensure continuity of the solution. The initial condition $\mathbf{d}_0 = \mathbf{q}_0$ is given by the initial joint states of the robot, denoted as
$$\mathbf{s}_0 = (\mathbf{q}_0, \dot{\mathbf{q}}_0, \ddot{\mathbf{q}}_0).$$
Based on the joint-space waypoint list $\mathcal{J}$, the objective of the low-level motion planner is to generate a smooth joint-space trajectory function
$$\hat{\mathbf{s}}(t) = (\hat{\mathbf{q}}(t), \dot{\hat{\mathbf{q}}}(t), \ddot{\hat{\mathbf{q}}}(t)), \quad (2)$$
which defines the expected joint position, velocity, and acceleration at time $t$. The output of this function is evaluated at discrete time intervals defined by the control frequency $f_c$, and the corresponding reference is sent to the robot in real time.

Starting from the current joint states $\mathbf{s}_0$, The trajectory must satisfy the constraint that each interpolated curve passes through the target joint position $\mathbf{d}_i$, precisely at time $T_i$, and it must reach the final target $\mathbf{d}_N$ at time $T_N$ with both velocity and acceleration equal to zero, ensuring a stable stop. Additionally, the trajectory should be optimized to minimize overall jerk throughout the entire motion duration, thereby improving motion smoothness and reducing mechanical stress.

*B. n-Degree Polynomial Function*

Once inverse kinematics has been resolved, trajectory interpolation is performed independently for each joint, as there is no coupling between degrees of freedom. Therefore, for a single joint, the position waypoints with the corresponding time stamps and the initial joint state can be denoted as follows:
$$\mathcal{D} = \{d_1, d_2, \ldots, d_N\}, \mathcal{T} = \{T_1, T_2, \ldots, T_N\}, s_0 = (d_0, v_0, a_0).$$
We adopt an n-degree polynomial interpolation method, where the degree of the polynomial is denoted by $L$. For each segment between two consecutive waypoints, an order-$L$ polynomial is fitted. Thus, we have for the interval $t \in (T_{i-1}, T_i)$, the trajectory for segment $n$ can be formulated as
$$\hat{q}_i(t) = \mathbf{b}_L(t)\mathbf{p}_i^T \quad (3),$$
where
$$\mathbf{b}_L(t) = [1, t, t^2, \ldots, t^L], \mathbf{p}_i = [p_{i,0}, p_{i,1}, p_{i,2}, \ldots, p_{i,L}].$$
Here, $\mathbf{b}_L(t)$ represents the monomial basis vector that defines the polynomial structure, and $\mathbf{p}_n$ is the coefficient vector for the $i^{th}$ segment. $\mathbf{p}$ is the collection of $\mathbf{p}_n$, where
$$\mathbf{p} = [\mathbf{p}_1^T, \mathbf{p}_2^T, \ldots, \mathbf{p}_N^T]^T \in \mathbb{R}^{(L+1)N}.$$
Once $\mathbf{p}$ is determined, the entire trajectory function is fully defined.

Figure 3 illustrates an example of a single joint's position-time profile. Upon receiving a motion request, the planner initializes the trajectory from the current joint position $d_0$. Based on the joint-space waypoints given, the planner generates a continuous reference position trajectory function
$$\hat{q}(t) = \{\hat{q}_1(t), \hat{q}_2(t), \ldots, \hat{q}_N(t)\}.$$
Furthermore, when the polynomial degree $L \geq 4$, the trajectory function and its first three derivatives, can be compactly expressed as:
$$\hat{q}_i^{(k)}(t) = \mathbf{b}_L^{(k)}(t)\mathbf{p}_i^T, \quad k = 0,1,2,3 \quad (4),$$
where $\mathbf{b}_L^{(k)}$ denotes the $k^{th}$ derivative of the monomial basis vector. This formulation allows direct evaluation of trajectory derivatives at any time $t$ which is particularly useful for imposing boundary constraints and optimizing dynamic smoothness. A polynomial degree of at least 4 is required to guarantee sufficient degrees of freedom to satisfy position, velocity, and acceleration boundary conditions at both ends of the segment. Therefore
$$\dot{\hat{q}}(t) = \{\dot{\hat{q}}_1(t), \dot{\hat{q}}_2(t), \ldots, \dot{\hat{q}}_N(t)\},$$
$$\ddot{\hat{q}}(t) = \{\ddot{\hat{q}}_1(t), \ddot{\hat{q}}_2(t), \ldots, \ddot{\hat{q}}_N(t)\}.$$
Thus, we have the one-dimensional joint state reference as
$$\hat{s}(t) = (\hat{q}(t), \dot{\hat{q}}(t), \ddot{\hat{q}}(t)) \quad (5).$$

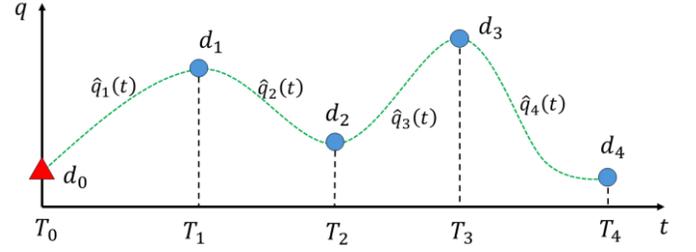

Fig. 3. Joint position over time for one joint. The red triangle indicates the current state at the time of request. The planner generates a continuous trajectory segments $\hat{q}_i(t)$, green dashed lines, based on the blue waypoints ($\mathbf{d}_i, T_i$).

*C. Trajectory Optimization*

The use of high-degree polynomial functions ensures that the joint-space trajectory function $\hat{q}_i(t)$ is inherently smooth. To further guarantee minimum overall jerk and ensure continuity of position, velocity, and acceleration between consecutive segments, as illustrated in Figure 3, the trajectory generation problem can be formulated as a **QP problem**,
$$\min_{\mathbf{p}_i} \sum_{i=1}^{N} \mathbf{p}_i^T \mathbf{Q}_i \mathbf{p}_i \quad (6).$$
Here, $\mathbf{Q}_i$ is the jerk cost matrix associated with the $i^{th}$ segment, defined as
$$\mathbf{Q}_i = \sum_{t \in \tau_i} \dddot{\mathbf{b}}_L(t)\dddot{\mathbf{b}}_L(t)^T \quad (7),$$
where $\tau_i = \{t_{i,j} \in [T_{i-1}, T_i]\}$ is a discrete set of time samples within the $i^{th}$ segment. The resolution of these time samples can be determined by arm control frequency, such that the number of sampling points per segment is equal to $f_c \cdot (T_{i-1} - T_i)$.

**Equality constraints** are imposed to ensure continuity and boundary conditions:

- Initial state constraints at $t = 0$:
$$\hat{q}_1(0) = q_0, \quad \dot{\hat{q}}_1(0) = \dot{q}_0, \quad \ddot{\hat{q}}_1(0) = \ddot{q}_0$$
- Terminal state constraints at $t = T_N$:
$$\hat{q}_N(T_N) = d_N, \quad \dot{\hat{q}}_N(T_N) = 0, \quad \ddot{\hat{q}}_N(T_N) = 0$$
- Continuity constraints at each segment junction $t = T_i$:
$$\hat{q}_i(T_i) = \hat{q}_{i+1}(T_i), \dot{\hat{q}}_i(T_i) = \dot{\hat{q}}_{i+1}(T_i), \ddot{\hat{q}}_i(T_i) = \ddot{\hat{q}}_{i+1}(T_i),$$
for $i = 1,2, \ldots, N-1$.

These constraints can be written in matrix form
$$\mathbf{A}_{eq}\mathbf{p} = \mathbf{b}_{eq} \quad (8),$$
which has
$$\mathbf{A}_{eq} \in \mathbb{R}^{(N+5) \times (L+1)N}, \mathbf{p} \in \mathbb{R}^{(L+1)N \times 1}, \mathbf{b}_{eq} \in \mathbb{R}^{(N+5) \times 1}$$

**Inequality constraints** are imposed to ensure physical feasibility:
$$|\dot{\hat{q}}_i(t_{i,j})| \leq v_{max}, \quad |\ddot{\hat{q}}_i(t_{i,j})| \leq a_{max}, \quad \forall t_{i,j} \in [T_{i-1}, T_i]$$
for $i = 1,2, \ldots, N$

These constraints can be written in matrix form
$$\mathbf{A}_{ineq}\mathbf{p} \leq \mathbf{b}_{ineq} \quad (9),$$
which has

$\mathbf{A}_{ineq} \in \mathbb{R}^{2 f_c T_N \times (L+1)N}, \mathbf{p} \in \mathbb{R}^{(L+1)N \times 1}, \mathbf{b}_{eqin} \in \mathbb{R}^{2 f_c T_N \times 1}$.

Finally, the trajectory generation problem can be formulated as a standard QP problem:
$$\min_{\mathbf{p}} \quad \mathbf{p}^T \mathbf{Q} \mathbf{p} \quad (10)$$
$$s.t. \quad \mathbf{A}_{eq}\mathbf{p} = \mathbf{b}_{eq}$$
$$\mathbf{A}_{ineq}\mathbf{p} \leq \mathbf{b}_{ineq},$$
where
$$\mathbf{Q} = blockdiag(\mathbf{Q}_1, \mathbf{Q}_2, \dots, \mathbf{Q}_N).$$

Once the request is received and the inputs $\mathcal{C}$, and initial state $\mathbf{s}_0$ is given, for each joint, the QP problem in Equation (10) can be constructed. Solving the QP yields the coefficient vector $\mathbf{p}$, which defines the trajectory $\hat{s}(t)$ in Equation (5). During real-time execution, the current time-from-start is passed into $\hat{s}(t)$ to compute the reference joint state for the controller.

This approach avoids precomputing the full trajectory, which may be computationally expensive or less accurate due to discretization. Instead, computing $\hat{s}(t)$ in real time ensures higher timing precision and system responsiveness.

### D. Interface Usage Modes

The proposed API can be used in three distinct usage modes, each designed for a different class of application scenarios. These modes share the same request format but differ in how the request is constructed and updated.

#### 1) Offline Planning

In this mode, the user provides a full Cartesian waypoint list $\mathcal{C}$. It is suitable for tasks requiring the end-effector to follow a specific path shape (e.g., a circle or straight line). To achieve higher fidelity to the desired path geometry, the user is advised to provide densely sampled waypoints such that the interpolated trajectory closely approximates the intended shape. This mode is typically used for precomputed motion sequences in structured environments.

#### 2) Interruptible Low-Frequency Control

This mode supports sending a single Cartesian waypoint at a time, where the current trajectory can be preempted by a new command before reaching the previous target. This behavior is particularly suited for AI agents that operate at low frequency and produce sparse or intermittent motion updates. By always executing only the most recent target, the system ensures both responsiveness and safety. If no further commands are received, the robot will smoothly decelerate and stop at the current goal.

#### 3) Teleoperation

In teleoperation scenarios, the master-side device (e.g., a VR controller) often only provides positional data without native velocity or acceleration measurements. Additionally, such devices typically operate at a lower frequency than the robot's control loop and may suffer from communication latency or jitter. To accommodate these constraints, the proposed API can be used as a buffered streaming interface: a fixed-size buffer is maintained, and at each master-side update cycle, a sequence of new waypoints is sent. In the buffered list $\mathcal{C} = \{\xi_1, \xi_2, \dots, \xi_N\}$, $\xi_1$ is the oldest point, and $\xi_N$ is the most recent. Durations are derived from master-side timestamps, and each new update can preempt the previous request. While this introduces a fixed delay, it can improve robustness under real-world teleoperation conditions.

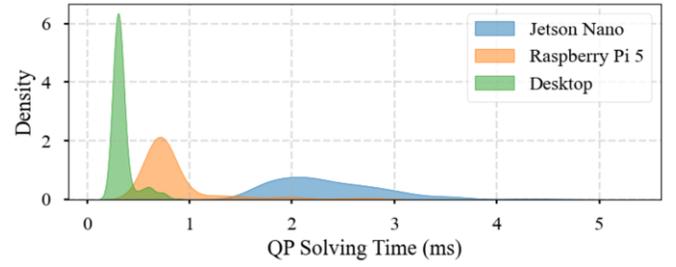

**Fig. 4.** QP Solving Time Distribution on Different Platforms Using OSQP

## IV. EXPERIMENTS

In this section, we first present a simple performance evaluation. Then, to validate the three API usage modes described in Section IV-C, we conduct three corresponding experiments, each reflecting a typical application scenario.

### A. Computational Performance Evaluation

Compared to solving the IK problem in formula (1) and the QP problem in formula (10), the time required to compute the reference state based on formula (5) is negligible. In specific manipulator configurations, the IK solver can be implemented analytically, significantly improving the real-time capability of generating a trajectory. For more general cases, fast numerical solvers such as TRAC-IK [24] can also provide efficient inverse kinematics solutions.

This experiment focuses on evaluating whether the formulated QP problem can be solved within the time constraints required for real-time motion planning. Specifically, the QP solving time must be shorter than $1/f_c$, the inverse of the control frequency, to ensure smooth trajectory updates without interruption from incoming requests.

We evaluate three platforms with different levels of computational capability: Jetson Nano with a quad-core 1.4 GHz CPU, Raspberry Pi 5 with a quad-core 2.4 GHz CPU, and a desktop equipped with a 12-core 4.5 GHz CPU. The OSQP solver [34] is used in all experiments and runs exclusively on the CPU. After tuning the solver parameters to ensure stable and optimal performance, we execute the full pipeline shown in Figure 2 on each device, controlling a 6-DOF collaborative robot arm in real time along a predefined trajectory. During each test, we send sequences of five Cartesian waypoints $\mathcal{C} = \{\xi_1, \xi_2, \dots, \xi_5\}$ to WOS at 20 Hz and collect 400 timing samples of the low-level motion planning node, which solves 6 parallel QP (one per joint) at $L = 5$.

The results are shown in Figure 4. All platforms achieve millisecond-level solving times. The desktop consistently achieves the fastest and most stable performance, with solving times typically around 0.5 ms. Despite its lower power consumption and smaller size, the Raspberry Pi 5 achieves an average solving time of approximately 0.7 ms, with a small number of outliers reaching up to 3 ms. The Jetson Nano, being the least powerful among the three, has a broader distribution centered around 2.1 ms, with some samples exceeding 5 ms. These findings suggest that the proposed algorithm can meet real-time control requirements at reasonably high frequencies on both high-performance desktop systems and compact modern embedded platforms.

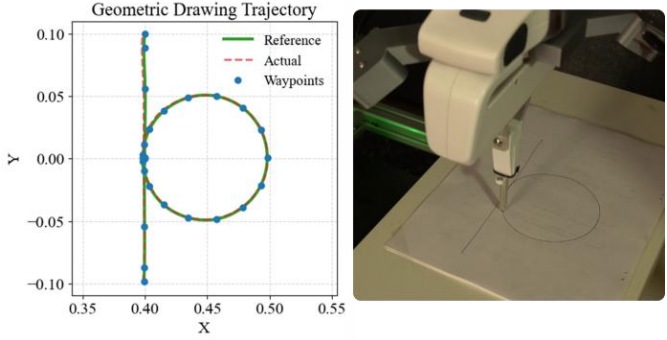

**Fig. 5.** Geometric Drawing Task: Planned Trajectory and Real Execution

All subsequent experiments are conducted on the same desktop as computing platform, using TRAC-IK for inverse kinematics and OSQP for QP solving, and all assigning $L = 5$.

### B. Offline Motion Planning: Geometric Drawing Tasks

This section aims to assess the applicability of the proposed real-time motion planning approach in offline scenarios and to examine whether the polynomial-based joint-space interpolation introduces any significant geometric distortion in Cartesian space.

The experiments were conducted on a Franka Research 3 (FR3) robotic arm equipped with a ballpoint pen at its end-effector. Multiple *rt-move-cartesian* requests were requests were issued, directing the arm to draw straight and circular trajectories for validation. Among them, one request specified a straight-line trajectory, and another described a circular trajectory. Each waypoint was assigned to a uniform duration of $D_i = 0.5s$. the straight-line trajectory consisted of 7 waypoints ($N = 7$), whereas the circular trajectory had 18 waypoints ($N = 18$). During execution, the robot controller operated at a fixed control frequency of $f_c = 100\ hz$.

As illustrated in Figure 5, the left plot presents the specified Cartesian waypoints $x_i, y_i$ the planned trajectory $\hat{\mathbf{s}}$ (green solid line), and the actual trajectory $\mathbf{s}$ (red dashed line), which was obtained via forward kinematics from the encoder data. The right plot shows a snapshot of the robot during execution. Figure 6 further depicts the evolution of Joint 7's state during task execution, focusing on two specific instances at $t = 5.5s$ and $t = 14.5s$. These snapshots illustrate both the completed and the upcoming segments of the trajectory, along with the corresponding encoder-based states.

Experimental results indicate that the robot was able to follow the prescribed Cartesian trajectories for both linear and circular drawing tasks with satisfactory accuracy, despite the use of polynomial interpolation in joint space. As illustrated in Figure 6, the joint trajectories remained smooth and continuously differentiable throughout the motion execution, thereby validating the temporal coherence of the proposed planning framework. While the interpolated paths may not achieve perfect geometric conformity at fine spatial scales, the resulting trajectory quality is sufficient for typical collaborative robotic applications such as welding, painting, and palletizing, where smoothness and task-level consistency are prioritized over micron-level precision.

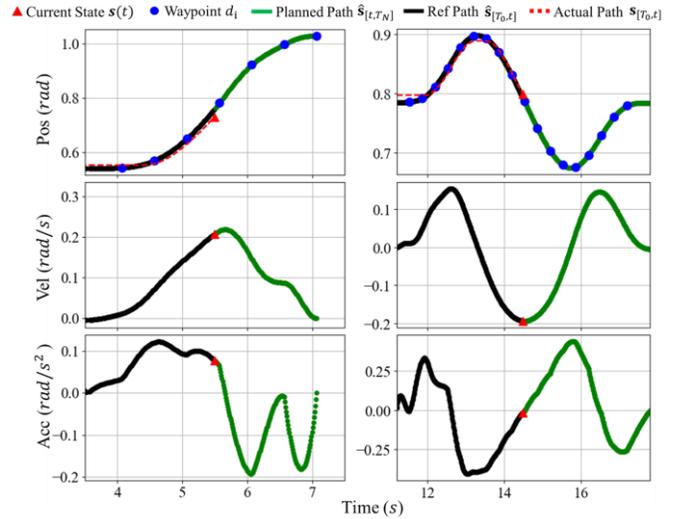

**Fig. 6.** Joint 7 trajectories during geometric drawing tasks. Left: straight-line drawing; right: halfway through circle drawing. Plots show planned trajectory $\hat{\mathbf{s}}$, waypoints set $\mathcal{D}$, and actual path $\mathbf{s}$ from encoder data.

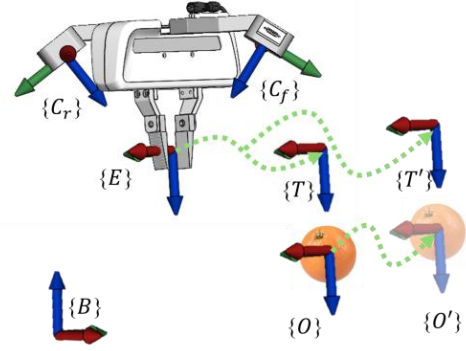

**Fig. 7.** System overview of stereo object localization and real-time interruptible planning for orange grasping.

### C. Interruptible Low-Frequency Replanning: Dynamic Object Grasping

#### 1) Dual-Camera Perception and 3D Object Localization

To evaluate the effectiveness of the proposed system under low-frequency interruptible motion planning, we designed a dynamic grasping scenario in which a mobile robot carries an orange while a fixed robotic manipulator (FR3) tracks and grasps the object when it stops moving.

Two Intel RealSense D405 depth cameras were mounted in front and behind, denoted as $\{C_f\}$ and $\{C_r\}$ as it shown in Figure 7, using the open-source CAD model [35], which provide stereo perception of the workspace. The dual-camera perception model estimates the orange's 3D position by leveraging instance masks from the SOLOv2 [36] detector. For each camera $\{C_f\}$ and $\{C_r\}$, the 3D point cloud corresponding to the segmented mask is extracted, and the average of the valid points yields the object center $\{O\}$ in each camera frame. Given the known rigid transformations between the end-effector frame $\{E\}$ and each camera frame $\{C_f\}$, $\{C_r\}$, we transform both estimates into the global robot base frame $\{B\}$. The final orange position is computed by averaging the two estimates in $\{B\}$, enhancing robustness to occlusion and noise.

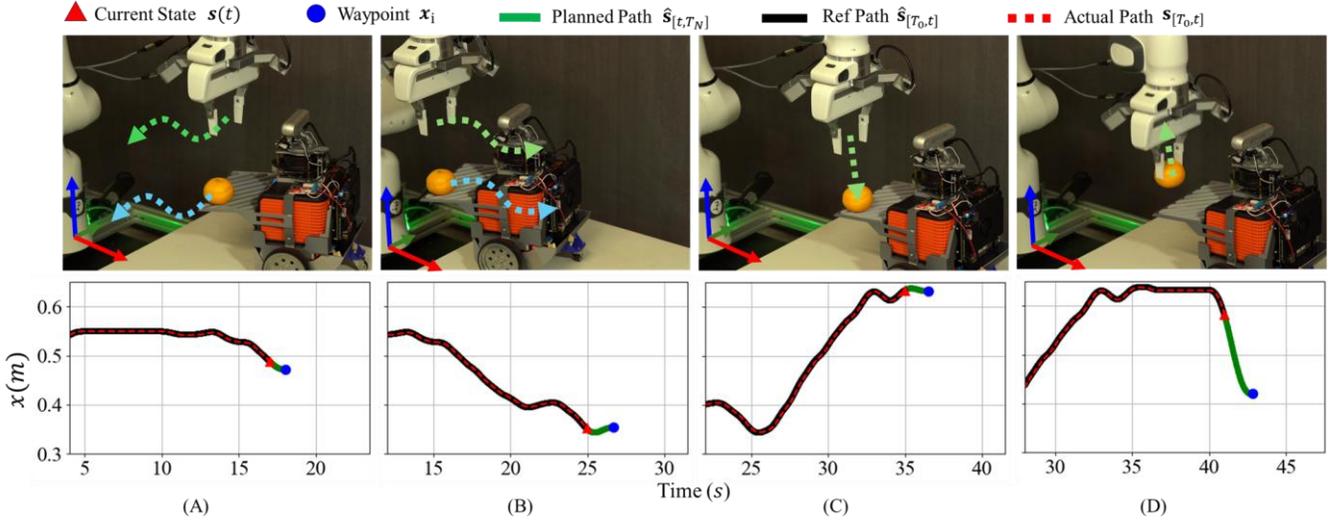

**Fig. 8.** Interruptible motion planning and grasping execution under dynamic target tracking using dual-camera perception.

*2) Interruptible Replanning and Grasp Triggering Logic*

As illustrated in Figure 7, the robotic arm continuously tracked the estimated target position. At each update step, upon observing the orange at position $\{O\}$, the robot issued a single waypoint *rt-move-cartesian* request to move toward the corresponding target pose $\{T\}$ (a location above the object) with a specified duration of 1.5 seconds. However, after 1 second into the execution, before reaching $\{T\}$, a new observation arrived at position $\{O'\}$, triggering an updated motion command toward a new target $\{T'\}$. The previous trajectory was immediately interrupted and replaced by the new plan. This process continued iteratively, allowing the robot to adapt to the dynamically changing target. Once the orange was detected to have stopped moving (i.e., displacement < 2 mm over consecutive cycles), the system triggered a grasping action, followed by a return-to-home sequence. The perception and target updating ran at a low frequency around 1 Hz, while the robot controller operated at a fixed control frequency of $f_c = 100\ hz$.

*3) Result Analysis*

The experiment demonstrates that the proposed system can reliably detect, track, and grasp dynamic objects under a low-frequency, interruptible planning paradigm. As illustrated in Figure 8, the robotic arm successfully follows the moving orange and executes a grasp once the object comes to a complete stop. The top row presents four key moments during the grasping sequence: A) The mobile robot moves in the negative x-direction of the global frame $\{B\}$. The orange is detected as moving, and the robot issues repeated *rt-move-cartesian* requests at about 1 Hz to interrupt ongoing trajectories, allowing the manipulator to refine its pose in the xy-plane toward the target. B) The mobile robot changes direction and starts moving along the positive x-axis, while the manipulator continues tracking. C) The object slows down and comes to a complete stop, triggering the grasp. D) The grasp is successfully completed. Green dashed arrows indicate the manipulator's trajectory, while blue dashed arrows indicate the motion of the mobile robot.

The bottom row shows the x-coordinate of the end-effector over time. The red curve represents the actual executed trajectory obtained via forward kinematics from encoder readings. The black segments represent completed planned trajectories, while green segments indicate newly issued plans. Blue dots mark the current active waypoint being tracked. The plot confirms that the system maintains trajectory continuity under a 1 Hz perception rate and responds promptly once the object becomes stationary, showcasing its robustness and real-time adaptability.

These results validate the applicability of interruptible motion planning strategies in dynamic and uncertain environments, highlighting the potential for mobile manipulation and human-robot collaboration applications.

D. *High-Frequency Real-Time Control: Dual-Arm Teleoperation*

*1) Experimental Setup*

This section presents a real-time teleoperation system involving two robotic manipulators: a human-operated Lite6 arm and an FR3 arm that mimics its motion in real time. The system is designed to complete a sequence of complex tasks remotely, including drawer opening, stamp retrieval, regrasping via a support bracket, stamping on a designated location, and returning the stamp to the drawer.

The operator interacts with a Delta6 force-sensing interface mounted on the Lite6 arm. The system reads force-induced displacements on the Delta6 interface and computes a virtual end-effector pose. This virtual pose is then mapped in real time to both the Lite6 and FR3 arms, enabling synchronized dual-arm motion.

*2) Teleoperation Logic*

A force-to-motion mapping is computed using the Delta6 model, and both the Lite6 and FR3 arms are driven simultaneously. The Lite6 arm is controlled using one waypoint *rt-move-cartesian API* at 50 Hz. Meanwhile, the FR3 arm receives *rt-move-cartesian* request at 25 Hz, each containing a buffer of $N = 5$ waypoints with uniform durations of $D_i =$ $0.04s$. Upon receipt, the FR3 controller executes the trajectory at a fixed control frequency of $f_c = 100\ hz$. Both arms are initialized to their calibrated home poses, and the end-effector transformations between Delta6, Lite6, and FR3 are defined by fixed offsets. Gripper control on FR3 is triggered through a keyboard interface.

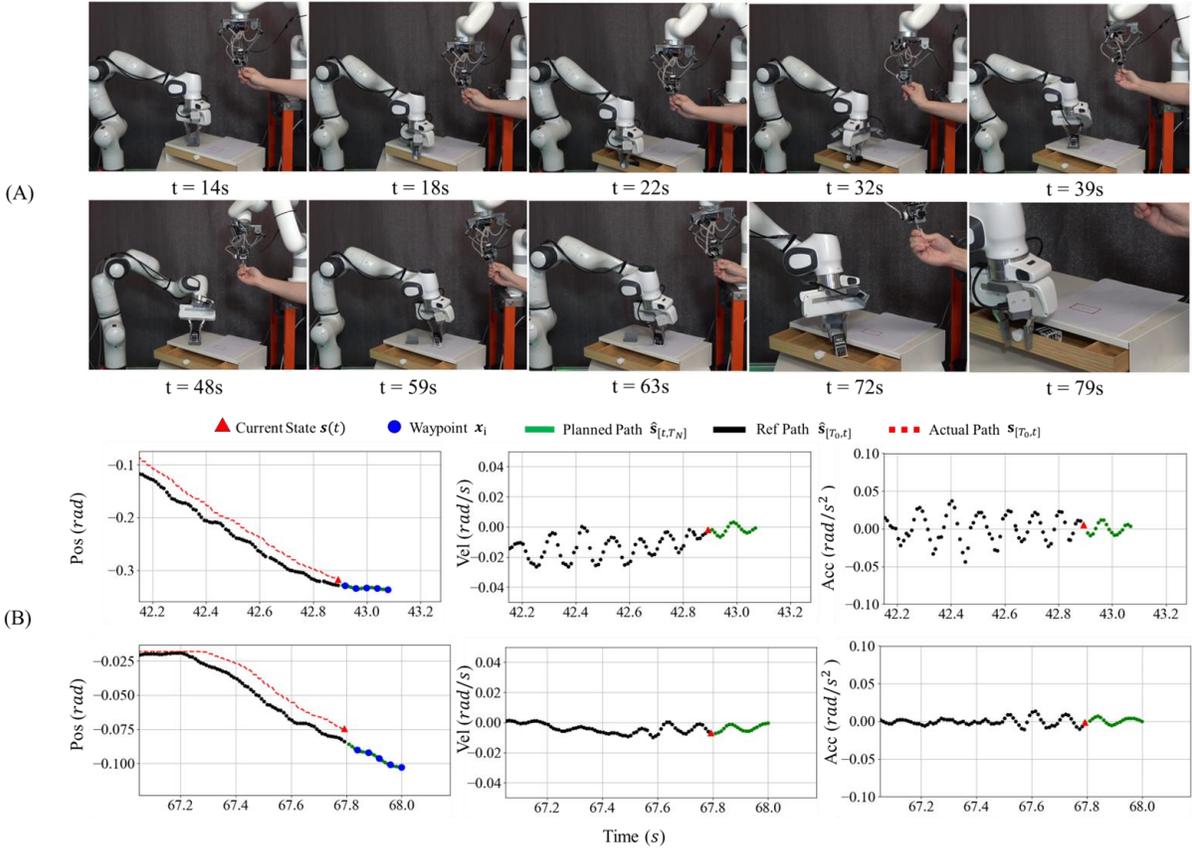

**Fig. 9.** Overview and trajectory analysis of the dual-arm teleoperation experiment. (A) Keyframes from the execution sequence, including drawer interaction, regrasping, stamping, and return motion. (B) Representative short-duration trajectories of Joint 3 around $t = 42.9s$ (dynamic phase) and $t = 67.8s$ (gentle phase), illustrating the continuity and smoothness of planned motion.

The teleoperation loop operates at 50 Hz and consists of the following steps: (1) Read encoder values from the Delta6 sensor and compute the current force-induced displacement. (2) Use forward kinematics to calculate the Lite6 end-effector pose. (3) Apply force-based control to compute the target pose for the Lite6 arm. (4) Send the target pose to Lite6 using *rt-move-cartesian*. (5) Every 2 cycles, compute the corresponding pose for the FR3 arm via rigid transformation and append it to a motion buffer. Once the buffer is filled, it is sent to FR3 using *rt-move-cartesian*. Gripper commands (open/close) are asynchronously triggered via keyboard events, allowing the user to directly control the gripper during teleoperation.

*3) Results and Observations*

Figure 9 shows the real-time operation of the dual-arm teleoperation system. Panel A highlights key stages of the task, where the FR3 arm smoothly follows the Lite6 arm's motion in a coordinated manner. Panel B presents two representative short-duration motion segments from Joint 4's state trajectory, corresponding to different motion intensities. The plots demonstrate that the proposed trajectory planning algorithm maintains continuous and smooth profiles for position, velocity, and acceleration across varying motion conditions. Notably, each planned segment naturally converges to zero velocity and acceleration at its endpoint, ensuring that in the absence of further incoming requests, the robot can still settle into a stable state smoothly.

These results validate the proposed motion planning framework's ability to support high-frequency, real-time teleoperation in unstructured environments. Although the buffer-based execution introduces a fixed delay of approximately 200 ms, it ensures temporal continuity and physical smoothness, key factors for safe and natural dual-arm collaboration.

## IV. CONCLUSION

This paper presents a unified, real-time low-level motion planning interface for collaborative robotic arms, designed for broad applicability across heterogeneous platforms. By integrating n-degree polynomial interpolation with a QP-based optimizer, the proposed method generates smooth, dynamically feasible joint-space trajectories from time-parameterized Cartesian waypoints. Through three representative experiments, offline drawing, interruptible grasping, and real-time dual-arm teleoperation, we demonstrate the interface's robustness, responsiveness, and generalizability. These results highlight the system's potential for deployment in structured, semi-structured, and dynamic environments alike.